\documentclass[letterpaper]{article} 
\usepackage{aaai23}  
\nocopyright
\usepackage{times}  
\usepackage{helvet}  
\usepackage{courier}  
\usepackage[hyphens]{url}  
\usepackage{graphicx} 
\usepackage{natbib}  
\usepackage{caption} 
\usepackage{algorithm}
\usepackage{algorithmic}
\usepackage{multirow}
\usepackage{array}
\newcolumntype{P}[1]{>{\centering\arraybackslash}p{#1}}
\newcommand{\cb}{\cellcolor{black!20}}
\usepackage{newfloat}
\usepackage{listings}
\DeclareCaptionStyle{ruled}{labelfont=normalfont,labelsep=colon,strut=off} 
\floatstyle{ruled}
\newfloat{listing}{tb}{lst}{}
\floatname{listing}{Listing}
\usepackage{amsmath,amsthm,amssymb}
\usepackage{comment}
\usepackage{enumitem}
\setlist[itemize]{leftmargin=*, noitemsep}

\usepackage{xcolor, colortbl}

\title{Uncertainty-Aware Reward-based Deep Reinforcement Learning for Intent Analysis of Social Media Information}
\author {
    Zhen Guo\textsuperscript{\rm 1}, 
    Qi Zhang\textsuperscript{\rm 1},
    Xinwei An\textsuperscript{\rm 2},
    Qisheng Zhang\textsuperscript{\rm 1},
    Audun J{\o}sang\textsuperscript{\rm 3},\\
    Lance M. Kaplan\textsuperscript{\rm 4},
    Feng Chen\textsuperscript{\rm 5},
    Dong H. Jeong\textsuperscript{\rm 6},
    Jin-Hee~Cho\textsuperscript{\rm 1}\\
    }
\affiliations {
    \textsuperscript{\rm 1}Department of Computer Science, Virginia Tech, VA, USA; 
    \textsuperscript{\rm 2}2810 Jackson Avenue, NY, USA;\\
    \textsuperscript{\rm 3}University of Oslo, Oslo, Norway; 
    \textsuperscript{\rm 4}DEVCOM Army Research Laboratory, MD, USA; \\   
    \textsuperscript{\rm 5}University of Texas at Dallas, Richardson TX, USA;
    \textsuperscript{\rm 6}University of the District of Columbia, DC, USA \\    
    zguo@vt.edu, qiz21@vt.edu, xan1@fordham.edu, qishengz19@vt.edu, audun.josang@mn.uio.no, lance.m.kaplan.civ@army.mil, feng.chen@utdallas.edu, djeong@udc.edu, jicho@vt.edu
} 

\begin{document}

\maketitle

\begin{abstract}
Due to various and serious adverse impacts of spreading fake news, it is often known that only people with malicious intent would propagate fake news.  However, it is not necessarily true based on social science studies.  Distinguishing the types of fake news spreaders based on their intent is critical because it will effectively guide how to intervene to mitigate the spread of fake news with different approaches.  To this end, we propose an intent classification framework that can best identify the correct intent of fake news.  We will leverage deep reinforcement learning (DRL) that can optimize the structural representation of each tweet by removing noisy words from the input sequence when appending an actor to the long short-term memory (LSTM) intent classifier.  Policy gradient DRL model (e.g., REINFORCE) can lead the actor to a higher delayed reward.  We also devise a new uncertainty-aware immediate reward using a subjective opinion that can explicitly deal with multidimensional uncertainty for effective decision-making.  Via  600K training episodes from a fake news tweets dataset with an annotated intent class, we evaluate the performance of uncertainty-aware reward in DRL.  Evaluation results demonstrate that our proposed framework efficiently reduces the number of selected words to maintain a high $95\%$ multi-class accuracy.  
\end{abstract}

\section{Introduction} \label{sec:intro}

Several recent studies~\cite{apuke2021fake, shen2021people} reported that fake news could be shared without bad intent, while it can contribute to a huge adverse impact on its propagation.  
Therefore, this work is motivated to provide a way of identifying the intent of fake news which can be used as the basis for effective intervention in mitigating fake news. More specifically, our intent classification method can contribute to identifying the right population and dealing with them differently to mitigate the impact of fake news propagation depending on their intent.   

Text classification tasks have been studied by deep learning features of embedding and structure representation~\cite{yogatama2017learning}.  Deep reinforcement learning (DRL) has been used to find an optimized structure representation by removing the noises, such as non-related words and lexical features~\cite{zhang2018learning}.  However, one of the main limitations in existing DRL-based text classification approaches is the use of a delayed reward to evaluate the prediction of the whole sequence of words.  Because the final processing state can only be determined when all words are processed, this hinders the learning process as a DRL agent cannot receive an immediate reward upon its local decision. 

This work tackles this issue by introducing a multidimensional uncertainty-aware reward in a DRL-based intent classifier. 
In addition, we propose an intent classification framework that can analyze the intents of fake and true news spreaders.  In addition, we consider multidimensional uncertainty estimates to identify optimal parameters. 



\vspace{1mm}
Our work makes the following {\bf key contributions}:
\vspace{-1mm}
\begin{enumerate}
\item We employ DRL algorithms to maximize the intent class prediction rate and minimize the number of words used in the intent classification process.  Specifically, we introduce an immediate multidimensional uncertainty-aware  reward to formulate an accumulated certainty reward. 
\item We use the uncertainty estimates, such as vacuity and dissonance, in updating the current policy.  This approach is the first to leverage a belief model, called {\em Subjective Logic}, that explicitly offers the capability of dealing with multidimensional uncertainty.
\item We resolve labor-intensive manual labeling tasks and annotate each news data either fake news or true news, by three annotators.  Finally, we assign the dominant intent class to each news data piece to combine the labels collected from three sources.
\end{enumerate}

\section{Related Work} \label{sec:related-work}


{\bf Intent Mining.}  
A number of social science research~\cite{ apuke2021fake, koohikamali2017information, shen2021people} 
has studied users' motivations for sharing fake news and their intent.   They found that the main reason for spreading fake news or false information was because they thought it was authentic and shared it unintentionally or even with good intent to help others or make fun.  Recent studies have proposed several natural language processing (NLP) methods to do intent mining in different task domains, such as user intent mining from reviews~\cite{khattak2021applying} and email intent~\cite{shu2020learning}.


{\bf DRL-based Text Mining.}  
DRL has been explored in NLP classification tasks to select meaningful word tokens and maintain the sequential relationship of selected words~\cite{zhang2018learning}.  That is, a better structure representation of whole text data can be learned by a DRL model to improve classification performance.  The sentiment classification research~\cite{jian2021aspect, wang2019aspect} 
has used DRL to reduce noisy tokens from whole sentence to improve the accuracy of a prediction model.  
Unlike the works using sentiment labels, our work identifies the intents of news articles user intention identification from user reviews and emails.  
Hence, we manually label the intent classes from an existing fake news dataset.  
To the best of our knowledge, this work is the first that considers multidimensional uncertainty in the immediate reward of actions and quantifies the multidimensional uncertainty through a belief model called {\em Subjective Logic} (SL)~\cite{Josang16-SL-book}.

\section{Intent Analysis Model} \label{sec:mudRIA}


\subsection{Intent Classes from Social Studies} \label{subsec:intent-classification}

In social sciences research, the intents of news spreaders, regardless of real or fake news, have been analyzed from the answers of questionnaires~\cite{koohikamali2017information, shen2021people}. However, no data-driven approaches have been used to study news spreaders' intents.  In our manual annotations of news articles, we considers the following five intent classes based on social sciences findings: 

\vspace{-1mm}
\begin{itemize}
\item {\bf Information sharing}: A common intent of online users' spreading behavior is purely sharing useful information to help other people.  This intent includes sense-making or expertise sharing to facilitate the truth of shared information~\cite{shen2021people}. 

\item {\bf Political campaign}: This intent uses fake news to falsely perceive an opponent party's political figure or group to mislead public opinions to win elections~\cite{purohit2019intent}. 

\item {\bf Socialization}: Online users share information to attract more friends and stay connected in online social networks (OSNs)~\cite{apuke2021fake}.  They can expand his/her social cycle by sharing news, often leading to finding common and interesting topics.  
\item {\bf Rumor propagation}: This intent misleads users by sharing a rumor or unverified information~\cite{purohit2019intent}.  Rumors can alter users' emotions and attitudes toward certain events and increase uncertainty. 

\item {\bf Emotion venting}: Online users can propagate fake news when they feel emotional happiness or disturbances (e.g., anger, depression, sadness) from reading good or bad events~\cite{alsmadi2021ontological}.  



\end{itemize}

\subsection{\bf Long Short-Term Memory (LSTM) Intent Classifier} \label{subsec:lstm}

Since the current dataset includes a collection of tweets, we call one piece of news data ``a news tweet'' for consistency.  First, we manually label each fake or true news tweet with an intent class in the Experiment Setup section.  Accordingly, this label is a gold intent class $\mathbf{y}$ for a news tweet $\mathbf{x}$.  Then, a classifier is learned from the annotated dataset to predict one's intent.  

\subsubsection{\bf Model Components.} This language model has a recurrent structure of $k$ embedding layers ($\theta_{em}$) and $k$ LSTM cells ($\theta_{lstm}$) to process a sequence of $k$ input words in a news tweet $\mathbf{x} = \{x_1, \ldots, x_k\}$.  After $k$ iterations with the final state $h_k$, a critic network ($\theta_{critic}$) with dense linear layers and ReLU and softmax activation functions can predict a distribution of $P$ intent classes.  We borrow the term `critic' to highlight the role in the DRL model.   
As in a traditional LSTM cell at iteration $t$, the inputs are a cell vector $c_{t-1}$, a hidden state vector $h_{t-1}$, and the input word vector $w_t$ embedded from a word token $x_t$.  The outputs are new $c_t$ and $h_t$ sent to the next iteration LSTM cell.  

\subsubsection{\bf Loss Function.}
A parameter set $\theta_{IC}$ of the three components in the intent classifier is updated from the training step, which minimizes the {\em cross-entropy loss} with the known gold intent labels.  Considering the over-fitting prevention, the L2 regularization term $||\theta_{IC}||_2^2$ is added to this loss by:
\begin{equation}\label{eq:loss-lstm}
\mathcal{L}_{IC} =  - \sum_{y=1}^P \hat{p}(y,\mathbf{x}) \log p(y|\mathbf{x}, \theta_{IC}) + \alpha ||\theta_{IC}||_2^2,
\end{equation} 
where $\alpha$ is the regularization rate, $p(y|\mathbf{x}, \theta_{IC})$ is predicted from the final softmax layer, and $\hat{p}(y,\mathbf{x})$ is the one-hot distribution of the gold label $\mathbf{y}$, which has $P$ values with a single high as 1 and all others are 0.

\subsection{DRL-based Intent Classifier} \label{subsec:drl}

\begin{figure}[t]
\centering
\includegraphics[width=0.48\textwidth]{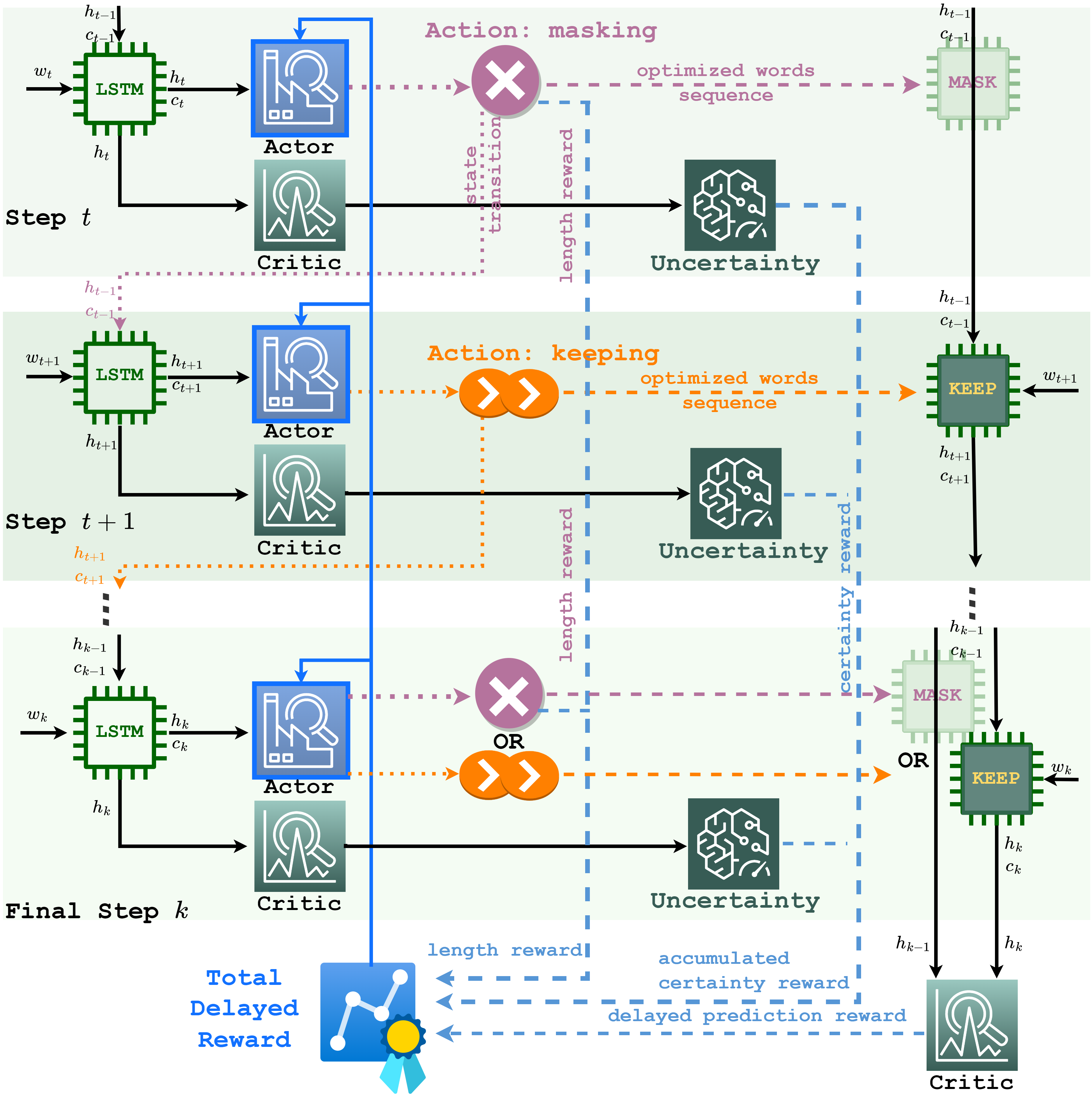}
\caption{The overview of the DRL model with an uncertainty-based immediate reward.} 
\label{fig:drl-framework}
\vspace{-3mm}
\end{figure}

This section discusses the sentence structure optimization steps by our policy gradient-based DRL (see Figure~\ref{fig:drl-framework} for detail).  Keeping the temporal relationship of the intent-related words can improve the intent prediction from the LSTM intent classifier described in the previous section.  Since the previous LSTM classifier serves as a fixed environment to support the state transitions, their parameters $\theta_{IC}$ are all frozen during the DRL learning of $\theta$.  

\subsubsection{Model Steps.} \label{subsubsec:drl-steps}

The DRL model adds a word selection step to the LSTM classifier to form a time series of actions $\tau = \{a_1,\ldots,a_k\}$ from each `keeping' or `masking' decision of the actor network.  Then, the selected or non-masked words generate an optimized input sequence $\mathbf{x'}=\{x_1, x_2, \dots, x_{k'}\}$ of $k'$ words ($k' \leq k$).  Finally, this shorter $\mathbf{x'}$ is processed by the three components of the LSTM intent classifier to predict the gold intent with higher accuracy.  
A training episode in DRL stands for one trial of the Markov decision process (MDP) by masking non-intent words from steps $1$ to $k$ of a news tweet.  In addition, an episode is applied based on only one news tweet data $\mathbf{x}$.  It is obvious that the golden class $\mathbf{y}$ for tweet $\mathbf{x}$ can be achieved during our annotation step.  The truth intent class $\mathbf{y}$ (e.g., socialization) can provide a delayed reward to lead the DRL to a higher reward by repeating five episodes in one training epoch.

\subsubsection{Word Selection DRL.} \label{subsubsec:wmm}
By setting $\mathbf{x'} = \mathbf{x}$, the DRL model extends an actor network to each LSTM cell $\theta_{lstm}$ in the previous LSTM intent classifier with the following details:  

\begin{itemize}
\item {\bf State}: At each step $t$, an input feature vector $w_t$ from the embedding layer $\theta_{em}$ represents the tokens of an input word $x_t$.  The LSTM cell $\theta_{lstm}$ processes $w_t$ with $c_{t-1}$ and $h_{t-1}$ and passes $c_t$ and $h_t$.  This hidden state vector $h_t$ serves as the state $s_t$ for the actor network at the current step as $s_t=\pi(w_t, c_{t-1}, h_{t-1}; \theta_{lstm})$. 

\item {\bf Actor}: The actor network has one dense linear layer followed by a ReLU activation and another dense linear layer followed by softmax activation of two nodes.  The parameters of all the neurons are in $\theta$ and can be trained by the DRL.  Given a state $s_t$, the actor generates a policy $\pi(a_t|s_t,\theta)$ for two actions.  

\item {\bf Action}: If an actor chooses $a_t=1$, the `keeping' action maintains $x_t$ in the optimized input sequence $\mathbf{x'}$.  Otherwise, if the actor chooses $a_t=0$, the `masking' action removes $x_t$ from $\mathbf{x'}$.

\item {\bf State Transition}:  The state transition determines how each action $a_t$ controls the state $s_{t+1}$ for the next step.  Since $s_{t+1}$ is generated from the next LSTM cell, $a_t$ decides the inputs for the LSTM cell of step $t+1$ by:
\begin{gather}
\label{eq:state-transition}
s_{t+1} = 
\begin{cases}
\pi(w_{t+1}, c_{t-1}, h_{t-1}; \theta_{lstm})\; \; \text{if $a_t=0$ `masking'};\\
\pi(w_{t+1}, c_{t}, h_{t}; \theta_{lstm})\; \; \; \;  \text{if $a_t=1$ `keeping'},
\end{cases}\raisetag{12pt}
\end{gather}
\end{itemize}

\subsubsection{Delayed Reward.} \label{subsubsec:delayed-reward}
A {\em delayed reward} is from the intent class distribution of the whole optimized input sequence $\mathbf{x'}$ of one news tweet, which is calculated from the LSTM classifier after the last word $w_{k'}$.  This delayed reward is a unique feature in some NLP problems when common immediate reward is unavailable.  Considering the prediction of the true intent class label, as $R_{pred}=p_{\theta_{critic}}(\mathbf{y}|\mathbf{x'})$, and the masked length reward of input words, the delayed reward is defined by:
\begin{equation}\label{eq:reward}
R = p_{\theta_{critic}}(\mathbf{y}|\mathbf{x'}) + \lambda (k-k')/k,
\end{equation}    
where $\theta_{critic}$ is a parameter for the critic network in the LSTM intent classifier, $\textbf{y}$ is the golden label for a news tweet $\textbf{x}$, $k'$ is the length of selected words in $\mathbf{x'}$, and $\lambda$ is a weight.

\subsubsection{\bf Policy Gradient.} \label{subsubsec:pg}
The goal of our DRL model is to maximize the delayed reward in Eq.~\eqref{eq:reward}.  Similar to many other NLP problems where an immediate reward is unavailable after deciding an individual step’s action, we will learn the actor's policy by policy gradient (PG)-based models.
For example, the current stochastic policy maximizes the expected delayed reward by calculating the gradients $R\log \pi(a_t|s_t,\theta)$ in REINFORCE algorithm~\cite{williams1992simple}.  
Based on the gradients for all $k$ transition steps in one episode following PG's on-policy property, REINFORCE calculates the {\em negative log loss} of the actor $\theta$'s policy as:
\begin{equation}\label{eq:loss-reinforce} 
\mathcal{L} = - \sum_t R\log \pi(a_t|s_t,\theta),
\end{equation}



\subsection{Uncertainty-based Immediate Reward} \label{subsec:uncertainty}




In each local step $t$ of the DRL model, when the hidden vector $h_t=s_t$ is passed directly to the critic network $\theta_{critic}$, the intent class distribution $p(s_t, \theta_{critic})$ from the critic network can be represented by Subjective Logic~\cite{Josang16-SL-book}.  Owing to the uncertainty metrics in the SL-based opinion, we can access the accumulated uncertainties or certainties from each $k$ decision steps in one training episode.     



\subsubsection{Formulating SL-based opinion from local intent classification probabilities.}\label{subsubsec:sl-opinion}

Remind that we adopt SL to consider multidimensional uncertainty where a traditional Dirichlet probability density function (PDF) can be easily mapped to a multinomial opinion which can provide a way to estimate different types of uncertainties.  
These local intent probabilities $p(s_t, \theta_{critic})$ can be regarded as a multinomial opinion in SL corresponding to five beliefs towards the intent classes. The SL opinion considers uncertainty metrics, such as {\em vacuity} caused by a lack of evidence and {\em dissonance} introduced by conflicting evidence.  Since the local intent probabilities cannot reflect the level of vacuity, we apply the vacuity maximization technique~\cite{Josang16-SL-book} on the local probabilities of $P$ classes and generate a vacuity maximized opinion $\omega_t = [\mathbf{b_t}$, $\mathbf{u_t}$, $\mathbf{g}]$.  There are $P$ belief masses in SL opinion as $\mathbf{b_t}$.  The $\mathbf{u_t} = [u^{vac}_t, u^{dis}_t]$ refers to two considered uncertainty metrics.  The base rates in vector $\mathbf{g}$ are the distribution of $P$ classes in annotated set $\mathcal{D}$.

\subsubsection{Total Delayed Reward based on Accumulated Certainty Estimates.}
Based on the previous discussion of the critic's immediate certainty metrics collected from local steps, the accumulated certainty metric serves as a new component of the total delayed reward by extending Eq.~\eqref{eq:reward} as:
\begin{equation}\label{eq:reward-delay}
R_{delay} = p_{\theta_{critic}}(\mathbf{y}|\mathbf{x'}) + \lambda (k-k')/k + \beta \sum_t (1-u_t).
\end{equation}  
where $u_t$ is either vacuity $u^{vac}_t$ or dissonance $u^{dis}_t$, and $\beta$ is the weight of accumulated certainty reward.  


\section{Experiment Setup} \label{sec:ex-setup}


\subsubsection{Datasets.} \label{subsec:datasets} We use the publicly available dataset {\em LIAR 2015}~\cite{wang2017liar}, which has 2,511 fake and 2,073 real news tweets verified by fact-checking agencies. 

\subsubsection{Intent annotation.}\label{subsubsec:annotation}  
Each news tweet is manually annotated by three annotators to give one of the five intent classes discussed.  Then, $33\%$ of all news is annotated as a dataset $\mathcal{D}$ of 1,500 tweets, with 835 fake and 665 true news.  Finally, we assign the dominant intent class from the three sources to each news tweet.  The base rate $\mathbf{g}$ for five classes are: `information sharing': 0.423, `political campaign': 0.275, `socialization': 0.135, `rumor propagation': 0.086, and `emotion venting:' 0.081.  
\subsubsection{Data preparation.}\label{subsubsec:data-preparation}  The annotated dataset $\mathcal{D}$ is split into fake news only, true news only, and both.  By training LSTM and DRL models, we use $80\%$ as a training set and $20\%$ as a testing set.  We use prefix padding of length $k=20$.     

\subsubsection{Parameterization.} \label{subsec:settings}

Our proposed DRL model aim to improve intent prediction by optimizing sentence representation structures, which will cause a reduction of words from the input data of the LSTM intent classifier.  In addition, we discuss the role of uncertainty-aware rewards by the SL in intent prediction.  Table~\ref{tab:params} summarizes the model hyper-parameters and their default settings.

\begin{table}[t]
\scriptsize
    \caption{\sc \centering Model Hyper-parameters Setting}
    \label{tab:params}
    \vspace{-2mm}
    \centering
    \begin{tabular}{|p{0.7cm}|p{1.7cm}|P{0.5cm}||p{2.45cm}|P{0.9cm}|}
    \hline
    \bf Model &\bf Parameter &\bf Value &\bf Parameter &\bf Value\\\hline
\multirow{4}{*}{LSTM} & Training epochs & 15 & Padding words length $k$ & 20 \\\cline{2-5}
    & Learning rate & 0.0003 & Dimensions of $h_t$ and $c_t$ & 128, 128 \\\cline{2-5}
    & Dropout rate & 0.25 & Dense layers dimensions & [257, 5] \\\cline{2-5} 
    & Batch size & 32 & Embedding dimensions& 128 \\\cline{1-5} 
\multirow{2}{*}{DRL} & Training epochs& 100 & Dense layers dimensions  & [257, 2]\\\cline{2-5}
    & Learning rate & 0.01 & Mini-batch episodes & 5\\\cline{2-5}
    & Batch size & 32 & Length reward weight $\lambda$ & 0.5 \\\hline 
    \end{tabular}
    \vspace{-3mm}
\end{table}

\begin{itemize}
\item LSTM non-RL model (named `LSTM') 
learns the parameters $\theta_C$ 
based on the loss function in Eq.~\eqref{eq:loss-lstm}.  

\item REINFORCE (named `DRL', from section DRL-based Intent Classifier section) is a basic PG method based on the delayed reward Eq.~\eqref{eq:reward} and loss function Eq.~\eqref{eq:loss-reinforce}.  

\item REINFORCE with uncertainty-aware reward (named `DRL-CV' and `DRL-CD' for vacuity and dissonance)  
uses the loss function in Eq.~\eqref{eq:loss-reinforce} by replacing $R$ with Eq.~\eqref{eq:reward-delay}.  
\end{itemize}



\subsubsection{Metrics.} \label{subsec:metrics}

Our DRL framework is evaluated by measuring (1) the multi-class classification accuracy counted by the ratio of correctly predicted data over the total of the testing data; (2) the effectiveness of DRL by prediction of the gold intent class from the critic network; and (3) the efficiency for DRL represented by the length of the optimized sequence.

\section{Experiment Results \& Analysis} \label{sec:results}
During one DRL training epoch of each news tweet, we collect the states, actions, and rewards by running five mini-batch episodes.  Thus, each epoch contains $6,000$ training episodes.  Our models can converge within 100 epochs and there are a maximum of 600K episodes.  In the testing of DRL models, the actors always follow the policy strictly.

\subsection{Multi-class Classification Accuracy}
Although this accuracy is not directly optimized in the DRL algorithm and rewards, it is crucial for our intent classification goal.  We list the accuracy scores for five classes and each intent class, along with the gold class prediction $R_{pred}$ in Table~\ref{tab:accuracy}.  The highest accuracy scores for DRL models with different weights $\lambda$ are within the same range, so we only show the case of $\lambda=0.5$ to compare to the DRL with certainty reward models.  The individual intent classes in the last column follow the same order in $\mathbf{g}$.  
The basic DRL model improved the gold class prediction by $5.3\%$ and increased the intent class classification by $11\%$ for all news data.  The LSTM model has similar accuracy for fake and true news, but all DRL models classified fake news with higher accuracies than true news.  Although the overall accuracy under our DRL-CV and DRL-CD is similar, the new certainty reward shows a slight improvement for the annotated intent class prediction.  Also, in general comparison, the two uncertainty-based DRL models can increase the accuracy for classes 1 `information sharing' and 2 `political campaign' but decrease the prediction of classes 3 `socialization' and 4 `rumor propagation'.  

\begin{table}[t]
\scriptsize
    \caption{\sc \centering Multi-class Accuracy from Testing}
    \label{tab:accuracy}
    \vspace{-2mm}
    \centering
    \begin{tabular}{|P{0.98cm}|P{0.45cm}|P{1.15cm}|P{0.53cm}|P{3.2cm}|}
    \hline
    \bf Model ($\lambda, \beta$) &\bf Test Data &\bf Gold Class Prediction &\bf Acc. & \bf Acc. by Intent Class\\\hline
    
\multirow{3}{*}{LSTM} & Fake & 0.806 & 0.866  & [0.853, 0.852, 0.727, 0.941, 1.0] \\\cline{2-5}
     & True & 0.840 & 0.867 & [0.898, 0.727, 0.889, 0.714, 1.0]\\\cline{2-5}
     &\cb Total &\cb 0.820 &\cb 0.860 &\cb [0.871, 0.802, 0.792, 0.850, 1.0]\\\cline{1-5}
     
\multirow{3}{1.1cm}{DRL \\ ($0.5$, N/A)} & Fake & 0.863 & 0.978 & [0.941, 1.0, 1.0, 1.0, 1.0] \\\cline{2-5} 
    & True & 0.887  & 0.958 & [0.966, 0.909, 1.0, 0.857, 1.0]\\\cline{2-5}
    &\cb Total &\cb 0.873 &\cb 0.970 &\cb [0.951, 0.963, 1.0, 0.943, 1.0]\\\hline{}
     
\multirow{3}{1.2cm}{DRL-CV ($0.5, 0.01$)} & Fake & 0.871 & 0.983 & [0.956, 1.0, 1.0, 1.0, 1.0] \\\cline{2-5}
     & True & 0.886 & 0.950 & [0.966, 0.955, 0.944, 0.714, 1.0] \\\cline{2-5}
     &\cb Total &\cb 0.877 &\cb 0.970 &\cb [0.960, 0.982, 0.978, 0.886, 1.0]\\\hline
     
\multirow{3}{1.2cm}{DRL-CD ($0.5, 0.05$)} & Fake & 0.874 & 0.983 & [0.956, 1.0, 1.0, 1.0, 1.0]\\\cline{2-5}
     & True & 0.890 & 0.958 & [0.966, 0.955, 1.0, 0.714, 1.0] \\\cline{2-5}
     &\cb Total &\cb 0.880 &\cb 0.973 &\cb [0.960, 0.982, 1.0, 0.886, 1.0]  \\\hline     
    \end{tabular}
    \vspace{-5mm}
\end{table}

Basically, DRL models show a high multi-class accuracy of $97\%$ across all settings.  Our DRL models aim to maximize the prediction accuracy while reducing the length of the optimized sequences.  Hence, in the next sections, we use at least $95\%$ overall multi-class accuracy performance to check the achieved minimum length for each DRL model.

\subsection{Effectiveness Reward by Prediction}
\begin{figure}[t]
\centering
\includegraphics[width=0.4\textwidth, height=0.25\textwidth]{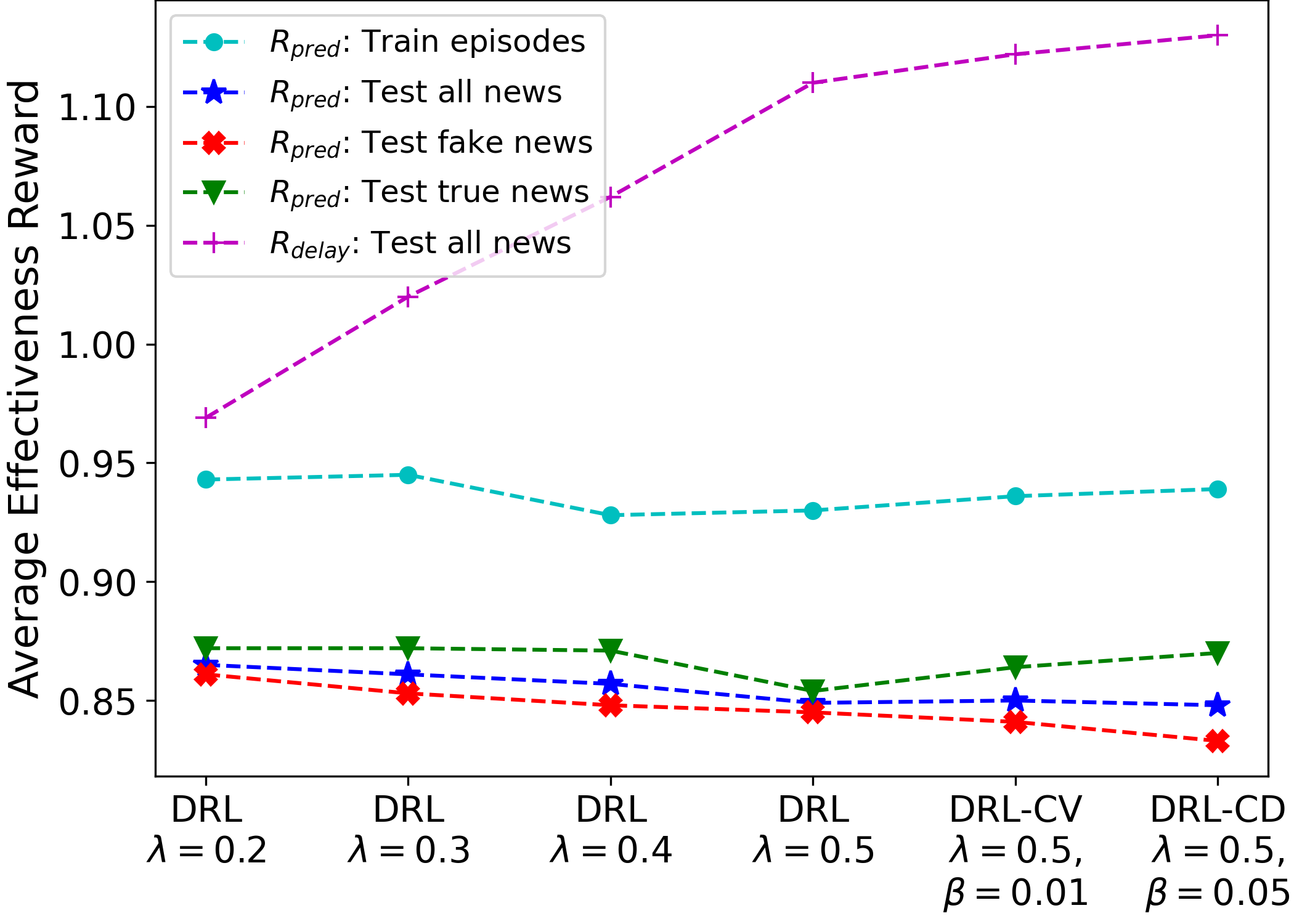}
\caption{Effectiveness scores from DRL models.} 
\label{fig:prediction}
\vspace{-3mm}
\end{figure}

Our reward function in Eq.~\eqref{eq:reward-delay} can maximize the effectiveness metric $R_{pred}$ for the direct gold class prediction.  We show this metric from the training episodes and the testing data, and compare it with the total delayed reward in Figure~\ref{fig:prediction}.   For the basic DRL models, when the weight $\lambda$ of the removed length is higher, both the training and testing effectivenesses decrease while the total reward still increases.  This means the loss of effectiveness is compensated by the length reward.  However, the uncertainty-related reward can increase training and testing effectiveness, compared to the DRL model with the same weight $\lambda=0.5$.  In addition, the effectiveness of true news data is improved with the help of the certainty reward.    

\subsection{Efficiency by Optimized Length}
\begin{figure}[t]
\centering
\includegraphics[width=0.4\textwidth, height=0.25\textwidth]{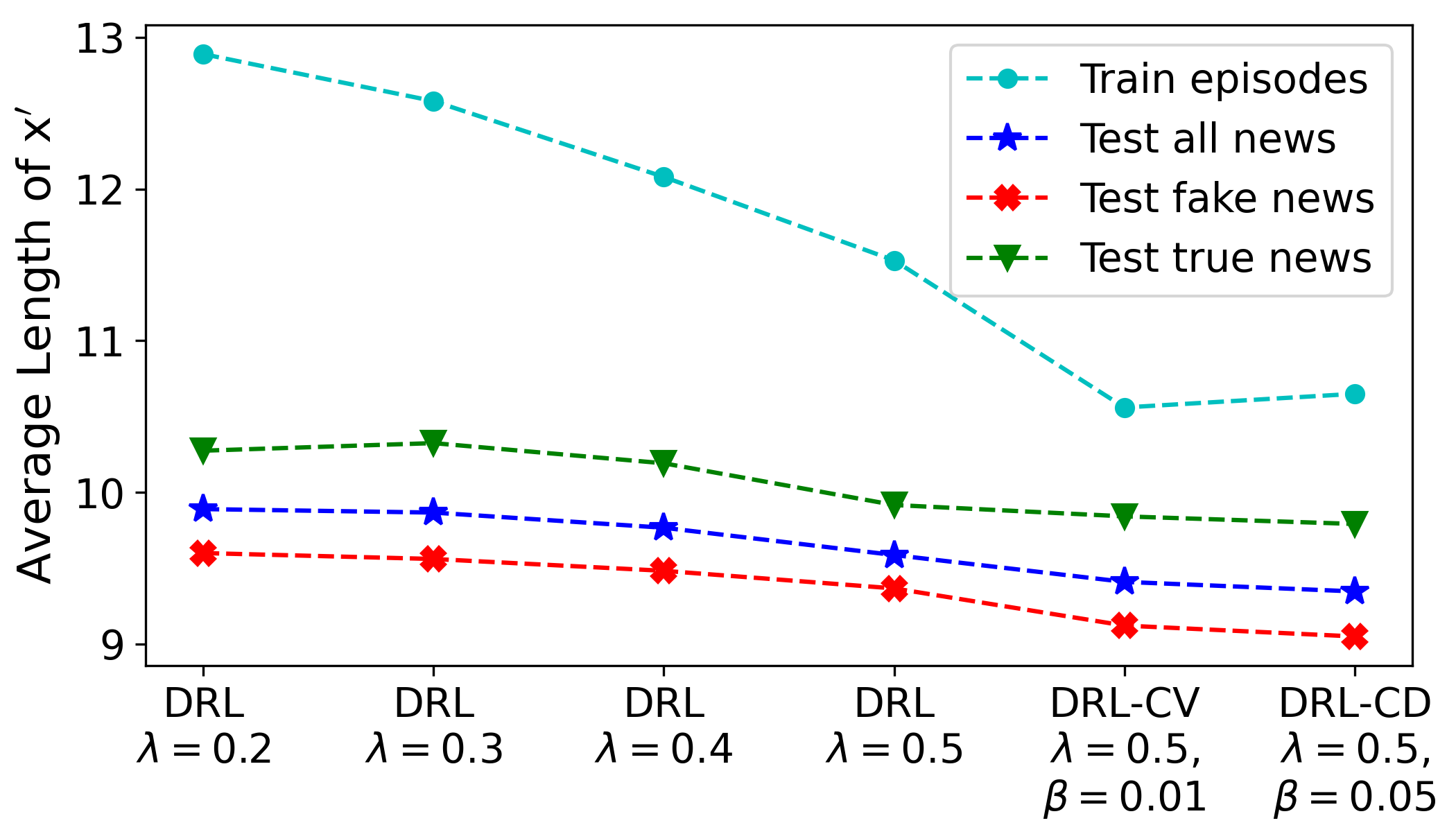}
\caption{Efficiency metric from DRL models.} 
\label{fig:length}
\vspace{-3mm}
\end{figure}

The masked length reward in Eq.~\eqref{eq:reward-delay} is DRL's main contribution to reducing the number of noisy words in the news tweets.  Figure~\ref{fig:length} illustrates the minimum number of words to maintain an overall classification accuracy of $95\%$.  Both training and testing lengths are reduced by adding a larger weight $\lambda$ for the masked length, meaning that the DRL models maintain $95\%$ classification accuracy while keeping less relevant words.  Fake news in testing removes more words than true news, leading to a higher reward.  The two uncertainty-related models reduce the length of both training and testing, indicating that the additional small amount of certainty values, both vacuity and dissonance, can help the DRL agent to explore more to the status of a shorter length.  The decrease of the kept length of DRL-CV and DRL-CD, compared to DRL with $\lambda=0.5$, can increase the efficiency reward and finally achieves a higher total delayed reward, as shown by the magenta line in Figure~\ref{fig:prediction}.     

\section{Conclusions} \label{sec:conclusion}

We annotated the intent class for the fake and true news from the dataset LIAR15.  We used this annotated dataset to train the LSTM intent classifier, and then created a DRL model to help reduce the noisy words.  The DRL model with optimized structure representation greatly improved the multi-class classification accuracy from the pretrained LSTM intent classifier. We then added an uncertainty-aware reward to help the DRL model reduce the length of words further, while maintaining the high level of intent class classification accuracy.  Our findings prove that two uncertainty metrics, vacuity and dissonance, can help the DRL agent's training to reach shorter lengths and higher the gold class predictions.  The two uncertainty-aware reward models can also decrease the length of testing data and increase the effectiveness of true news testing data, resulting in attaining a higher total delayed reward for the total test data.  We will dig into more relevant metrics for the rewards by effectiveness and efficiency in our future work. 

\section{Acknowledgments}
This work is partly supported by the Army Research Office under Grant Contract Number W91NF-20-2-0140 and NSF under Grant Numbers 2107449, 2107450, and 2107451.

\bibliography{ref}

\begin{thebibliography}{14}
\providecommand{\natexlab}[1]{#1}

\bibitem[{Alsmadi, Alazzam, and AlRamahi(2021)}]{alsmadi2021ontological}
Alsmadi, I.; Alazzam, I.; and AlRamahi, M.~A. 2021.
\newblock An ontological analysis of misinformation in online social networks.
\newblock \emph{arXiv:2102.11362}.

\bibitem[{Apuke and Omar(2021)}]{apuke2021fake}
Apuke, O.~D.; and Omar, B. 2021.
\newblock Fake news and COVID-19: modelling the predictors of fake news sharing
  among social media users.
\newblock \emph{Telematics and Informatics}, 56: 101475.

\bibitem[{Jian et~al.(2021)Jian, Nayak, Majumder, and Poria}]{jian2021aspect}
Jian, S. Y.~B.; Nayak, T.; Majumder, N.; and Poria, S. 2021.
\newblock Aspect sentiment triplet extraction using reinforcement learning.
\newblock In \emph{CIKM 2021}, 3603--3607.

\bibitem[{J{\o}sang(2016)}]{Josang16-SL-book}
J{\o}sang, A. 2016.
\newblock \emph{Subjective Logic: A Formalism for Reasoning Under Uncertainty}.
\newblock Springer.

\bibitem[{Khattak et~al.(2021)Khattak, Habib, Asghar, Subhan
  et~al.}]{khattak2021applying}
Khattak, A.; Habib, A.; Asghar, M.~Z.; Subhan, F.; et~al. 2021.
\newblock Applying deep neural networks for user intention identification.
\newblock \emph{Soft Computing.}, 25(3): 2191--2220.

\bibitem[{Koohikamali and Sidorova(2017)}]{koohikamali2017information}
Koohikamali, M.; and Sidorova, A. 2017.
\newblock Information Re-Sharing on Social Network Sites in the Age of Fake
  News.
\newblock \emph{Informing Science}, 20.

\bibitem[{Purohit and Pandey(2019)}]{purohit2019intent}
Purohit, H.; and Pandey, R. 2019.
\newblock Intent mining for the good, bad, and ugly use of social web:
  Concepts, methods, and challenges.
\newblock In \emph{Emerging Research Challenges and Opportunities in
  Computational Social Network Analysis and Mining}, 3--18. Springer.

\bibitem[{Shen et~al.(2021)Shen, Lee, Pan, and Lee}]{shen2021people}
Shen, Y.-C.; Lee, C.~T.; Pan, L.-Y.; and Lee, C.-Y. 2021.
\newblock Why people spread rumors on social media: Developing and validating a
  multi-attribute model of online rumor dissemination.
\newblock \emph{Online Information Review}.

\bibitem[{Shu et~al.(2020)Shu, Mukherjee, Zheng et~al.}]{shu2020learning}
Shu, K.; Mukherjee, S.; Zheng, G.; et~al. 2020.
\newblock Learning with weak supervision for email intent detection.
\newblock In \emph{ACM SIGIR 2020}, 1051--1060.

\bibitem[{Wang et~al.(2019)Wang, Zhou, Hu, and He}]{wang2019aspect}
Wang, T.; Zhou, J.; Hu, Q.~V.; and He, L. 2019.
\newblock Aspect-level sentiment classification with reinforcement learning.
\newblock In \emph{2019 IJCNN}, 1--8.

\bibitem[{Wang(2017)}]{wang2017liar}
Wang, W.~Y. 2017.
\newblock “Liar, Liar Pants on Fire”: A New Benchmark Dataset for Fake News
  Detection.
\newblock In \emph{ACL 2017 (Volume 2: Short Papers)}, 422--426.

\bibitem[{Williams(1992)}]{williams1992simple}
Williams, R.~J. 1992.
\newblock Simple statistical gradient-following algorithms for connectionist
  reinforcement learning.
\newblock \emph{Machine learning}, 8(3): 229--256.

\bibitem[{Yogatama et~al.(2017)Yogatama, Blunsom, Dyer, Grefenstette, and
  Ling}]{yogatama2017learning}
Yogatama, D.; Blunsom, P.; Dyer, C.; Grefenstette, E.; and Ling, W. 2017.
\newblock Learning to compose words into sentences with reinforcement learning.
\newblock In \emph{ICLR 2017}.

\bibitem[{Zhang, Huang, and Zhao(2018)}]{zhang2018learning}
Zhang, T.; Huang, M.; and Zhao, L. 2018.
\newblock Learning structured representation for text classification via
  reinforcement learning.
\newblock In \emph{Thirty-Second AAAI Conference}.

\end{thebibliography}

\end{document}